\documentclass[10pt,letterpaper]{article}

\usepackage{cogsci}
\usepackage{pslatex}
\usepackage{apacite}
\usepackage{tikz}
\usepackage{graphicx}
\usepackage{pgfplots}
\usepackage{makecell}
\usepackage{pifont}

\usepackage{tabu}
\usepackage{tabularx}
\usepackage{multirow}
\usepackage{booktabs}
\usepackage{array}

\usepackage{amsfonts}
\usepackage{mathtools}
\usepackage{bm}
\usepackage{url}
\usepackage{here}

\usepackage{amsmath}

\usepackage[hang,small,bf]{caption}
\usepackage[subrefformat=parens]{subcaption}
\usepackage{xcolor}
\definecolor{ggreen}{HTML}{9BBB59}
\definecolor{ppurple}{HTML}{9F4C7C}

\definecolor{bblue}{HTML}{88CCEE}
\definecolor{rred}{HTML}{CC6677}
\cogscifinalcopy % Uncomment this line for the final submission 

\usepackage[authoryear,round]{natbib}

\usepackage{pslatex}
\usepackage{float} % Roger Levy added this and changed figure/table
\title{How Panel Layouts Define Manga: Insights from Visual Ablation Experiments}

\author{
\textbf{Siyuan Feng} (\textbf{9445233883@g.ecc.u-tokyo.ac.jp})\textsuperscript{1}, 
\textbf{Teruya Yoshinaga} (\textbf{t-yoshinaga-3k6@eagle.sophia.ac.jp})\textsuperscript{2},\\ 
\textbf{Katsuhiko Hayashi} (\textbf{katsuhiko-hayashi@g.ecc.u-tokyo.ac.jp})\textsuperscript{1},\\
\textbf{Koki Washio} (\textbf{kkwashio3333@gmail.com})\textsuperscript{3}, 
\textbf{Hidetaka Kamigaito} (\textbf{kamigaito.h@is.naist.jp})\textsuperscript{4}
\\
\textsuperscript{1}The University of Tokyo, Department of Language and Information Sciences, Tokyo, Japan\\
\textsuperscript{2}Sophia University, Graduate Degree Program of Applied Data Sciences, Tokyo, Japan\\
\textsuperscript{3}Freelance Researcher, Tokyo, Japan\\
\textsuperscript{4}Nara Institute of Science and Technology, Division of Information Science, Ikoma, Nara, Japan
}

\begin{document}

\maketitle

\begin{abstract}

Manga has gained global popularity, yet how its visual elements, such as characters, text, and panel layouts, reflect the uniqueness of individual works remains underexplored. This study investigates the contribution of panel layouts to manga identity through both quantitative and qualitative analysis. We trained a deep learning model to classify manga titles based solely on facing page images, and performed ablation experiments by removing characters and text, retaining only panel frame structures. Using 10,122 images from 104 works in 12 genres in the Manga109 dataset, we demonstrate that panel layouts alone enable high-accuracy classification. Grad-CAM visualizations further reveal that the models focus on layout features such as size, spacing, and alignment. These findings suggest that panel layouts encode work-specific stylistic patterns and support visual narrative comprehension, highlighting their role as a key component of manga's visual identity.

\end{abstract}

\section{Introduction}

Comic is a globally popular medium, encompassing a wide range of themes and genres. With the increasing adoption of e-books, the digital comic market has seen significant growth, and many comic works are now available in electronic formats. This shift has drawn considerable attention to multimedia processing research for digital comics, particularly Japanese comics (referred to simply as ``manga" in this paper).

Manga, a distinctive form of Japanese comics, is characterized by its unique panel arrangements, which, along with creative character designs, guide the reader’s eye through the story in a carefully orchestrated manner. Panel layouts play a crucial role in conveying the narrative, as emphasized by Shogakukan’s “Newcomer Manga Award Manga Creator Training Course”\footnote{\url{https://shincomi.shogakukan.co.jp/training/004.html}}, which outlines five fundamental rules for panel layout design. Manga artists can further enhance originality by creatively adhering to or intentionally breaking these rules to suit their storytelling needs.
While readers may perceive subtle layout differences, it is often difficult to determine how these reflect the style of a work. This raises an important question: To what extent do panel layouts reflect the uniqueness of a manga work, and how do they contribute to its overall identity?

The significance of panel layouts has been acknowledged in multiple foundational works. For instance, \citet{cohn2013visual} emphasized the impact of page layout on reading order and compared the technical usage of panel arrangements between Japanese and Western comics. \citet{mccloud1993understanding} highlighted the role of panel-to-panel transitions and particularly explored how the spacing between panels (gutter) influences the flow of storytelling. Additionally, \citet{fusanosuke2020panel} delved deeply into the techniques and effects of panel layout in manga, offering insights into how layout decisions enhance the reader’s experience. While these works underscore the importance of panel layouts, they do not delve into the extent to which panel layouts contribute to defining the unique identity of a manga work.

Despite these foundational studies, existing research on manga has predominantly focused on characters, text, and other visual elements, with limited exploration of panel layouts. While some studies leverage engineering approaches, such as image processing or machine learning, these often focus on practical applications rather than delving into the artistic or structural characteristics of manga. This imbalance leaves the role of panel layouts in reflecting the creative uniqueness of manga largely unexplored.

\begin{figure}
    \centering
    \includegraphics[width=0.95\linewidth]{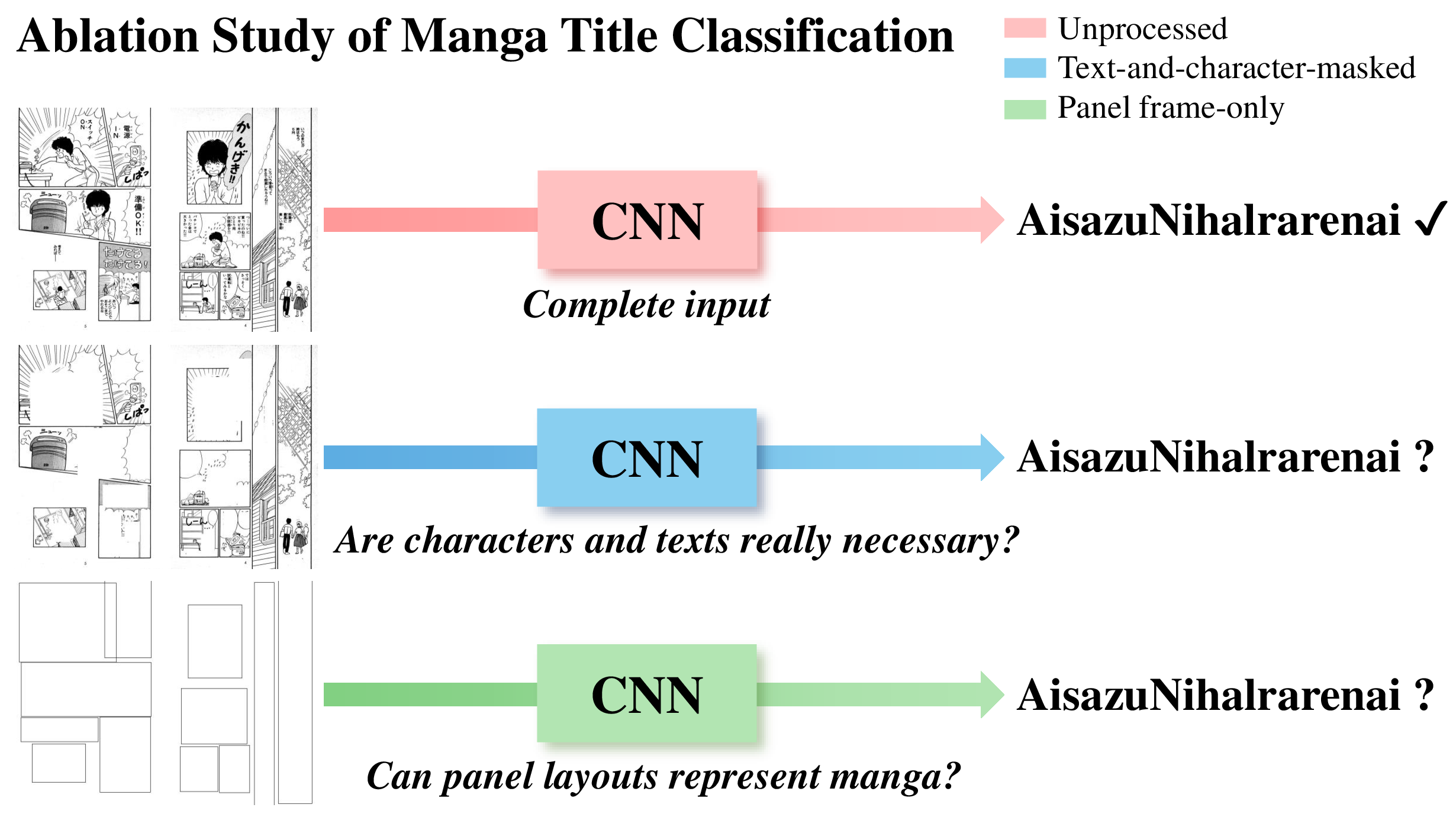}
    \caption{Brief description of the multi-class manga title classification experiment of the titles of the comics. The images shown in the figure are from AisazuNihalrarenai \copyright Masako Yoshi.}
    \label{fig:concept}
\end{figure}

To address this gap, in this paper, we conducted visual ablation experiments. The experimental process is shown in Figure~\ref{fig:concept}. We prepared three types of images: original unprocessed facing page images, text-and-character-masked facing page images, and panel frame-only facing page images. These images were used as input for a deep learning model to perform a multi-class classification task on manga titles.
The objective of this study was to analyze the changes in model performance by removing text and characters or retaining only panel layouts. If the model performs comparably well on processed images as it does on original unprocessed images, we believe this would validate the uniqueness of features such as manga panel layouts. By using panel frame-only facing page images, we aimed to evaluate the contribution of panel layout features from each work to the classification task.
Subsequently, we employed Grad-CAM~\citep{grad-cam} to visualize the key features in the input images and qualitatively interpret the results. Finally, we conducted additional experiments, including classification of manga publishers and genres, to rule out other factors influencing the uniqueness of panel layouts in manga works. We also performed classification experiments on noisy panel frames, further demonstrating that panel layouts are unique features of each manga work.

These results indicate that panel layouts possess inherent uniqueness, even in the absence of content such as characters or text. This suggests that layout structures carry stylistic information that may reflect the unique visual identity of each work. Although our approach is based on a classification task, the broader aim is to understand how layout alone encodes work-specific style and supports visual narrative interpretation. This line of inquiry is closely related to cognitive theories of attention and sequential image comprehension.

\section*{Related Work}

\subsection*{Visual Analysis in Creative Works}

Numerous studies have used images of creative works as input for predicting associated attributes such as titles or genres. \citet{chu-guo} fine-tuned a pretrained CNN to classify movie genres based on poster images. \citet{wi} applied a similar approach and visualized the model’s focus using Grad-CAM to verify whether it extracted expected features. \citet{godwin2018examination,takahashi2023textbook,tuchler2021impact} examined how layout affects comprehension or user experience, including children’s book design and web interfaces.

Several works also explored deep learning approaches in comic research. \citet{laubrock2020computational,augereau2018survey,sharma2024image} surveyed computer vision applications in comics. \citet{laubrock} conducted classification of comic authors and titles using CNNs on datasets like GNC~\citep{gnc} and Manga109~\citep{manga109,manga109-original}. \citet{balloon} performed speech balloon detection, while ~\citet{Rigaud} worked on text segmentation inside speech balloons. \citet{young2019feature} used CNN-based classification and feature visualization. \citet{kukreja2023unmasking} applied CNN and Random Forest models to classify manga pages.

\subsection*{Existing Issues in Manga Research}

While a growing number of studies have used manga page images, most focus on engineering applications, with limited attention to the underlying visual structure of manga. \citet{laubrock} attempted to analyze manga features through classification, but their method treated pages as undivided wholes, without isolating individual components like panel layout. \citet{young2019feature} proposed a CNN-based model for author classification and visualized learned features, but their analysis focused on stylistic textures and drawing patterns, rather than structural layout.

Although previous studies have explored page-level classification, none have specifically isolated panel layouts through ablation and visualization to analyze their contribution to manga identity. Although \citet{kukreja2023unmasking} trained a CNN to classify genres, they did not consider whether each manga has unique stylistic traits. Their dataset was small and did not evaluate features like panel layout independently. In contrast, our work focuses on isolating panel layout features through ablation and visual interpretation, to investigate their role in distinguishing manga works.

\section{Experiment}

In the experiment, we analyze the characteristics of manga works by training multi-class classification models to predict the title of the work using manga facing page images as input. Specifically, we extracted 10,122 images containing panel information out of 10,602 facing page images obtained from Manga109~\citep{manga109,manga109-original}, and classified each image into 104 works based on information from \url{http://www.manga109.org/en/index.html}. The breakdown of genres includes 12 categories: 4-panel, animal, battle, fantasy, history, horror, humor, love, romantic comedy, SF, sport, and suspense. Table~\ref{tab:datstats} provides an overview of the dataset.

\begin{figure}[htbp]
    \centering
    \begin{subfigure}[t]{0.15\textwidth} 
        \centering
        \includegraphics[width=\textwidth]{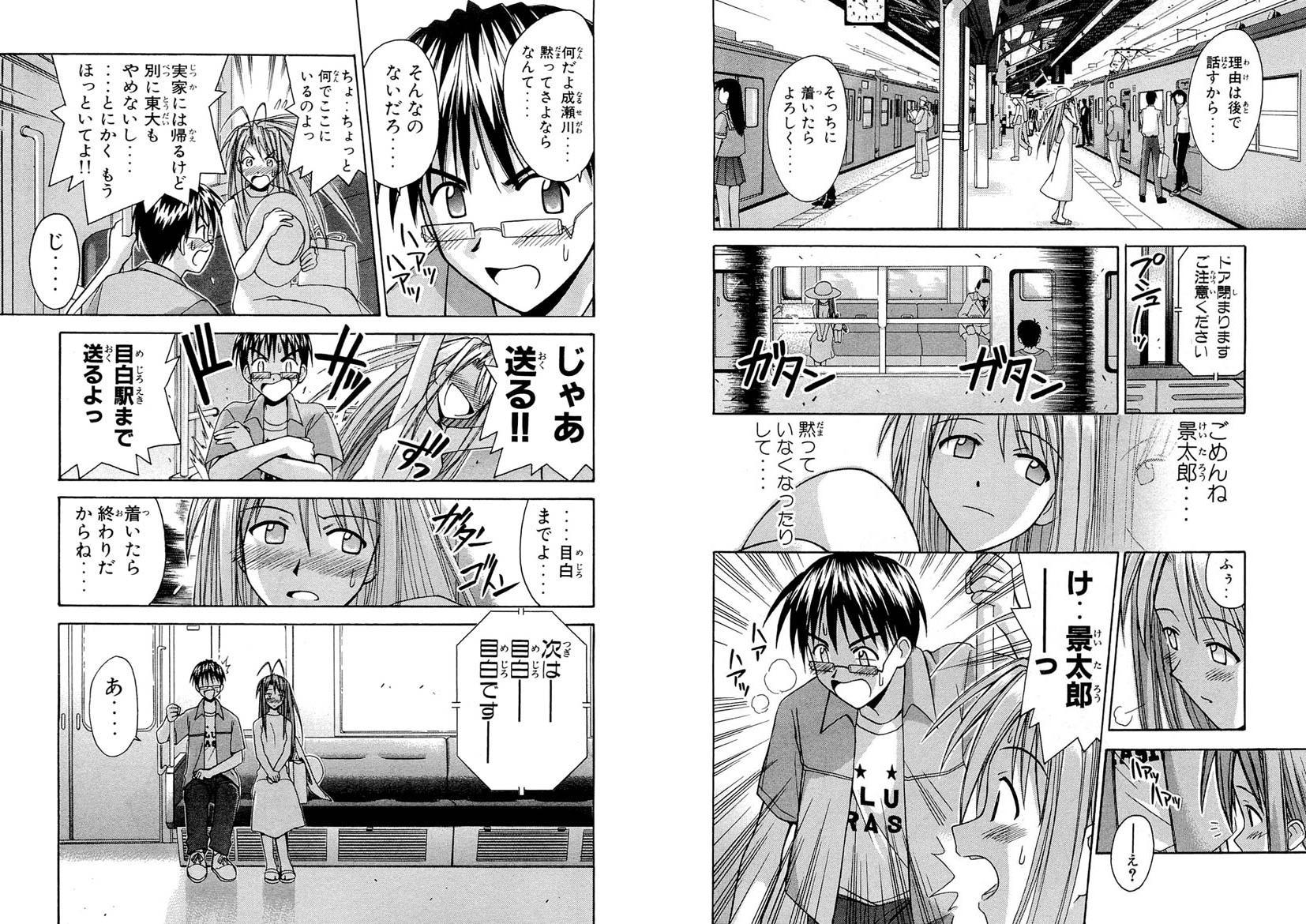} 
        \caption{Unprocessed} 
        \label{fig:a}
    \end{subfigure}
    \hfill
    \begin{subfigure}[t]{0.15\textwidth}
        \centering
        \includegraphics[width=\textwidth]{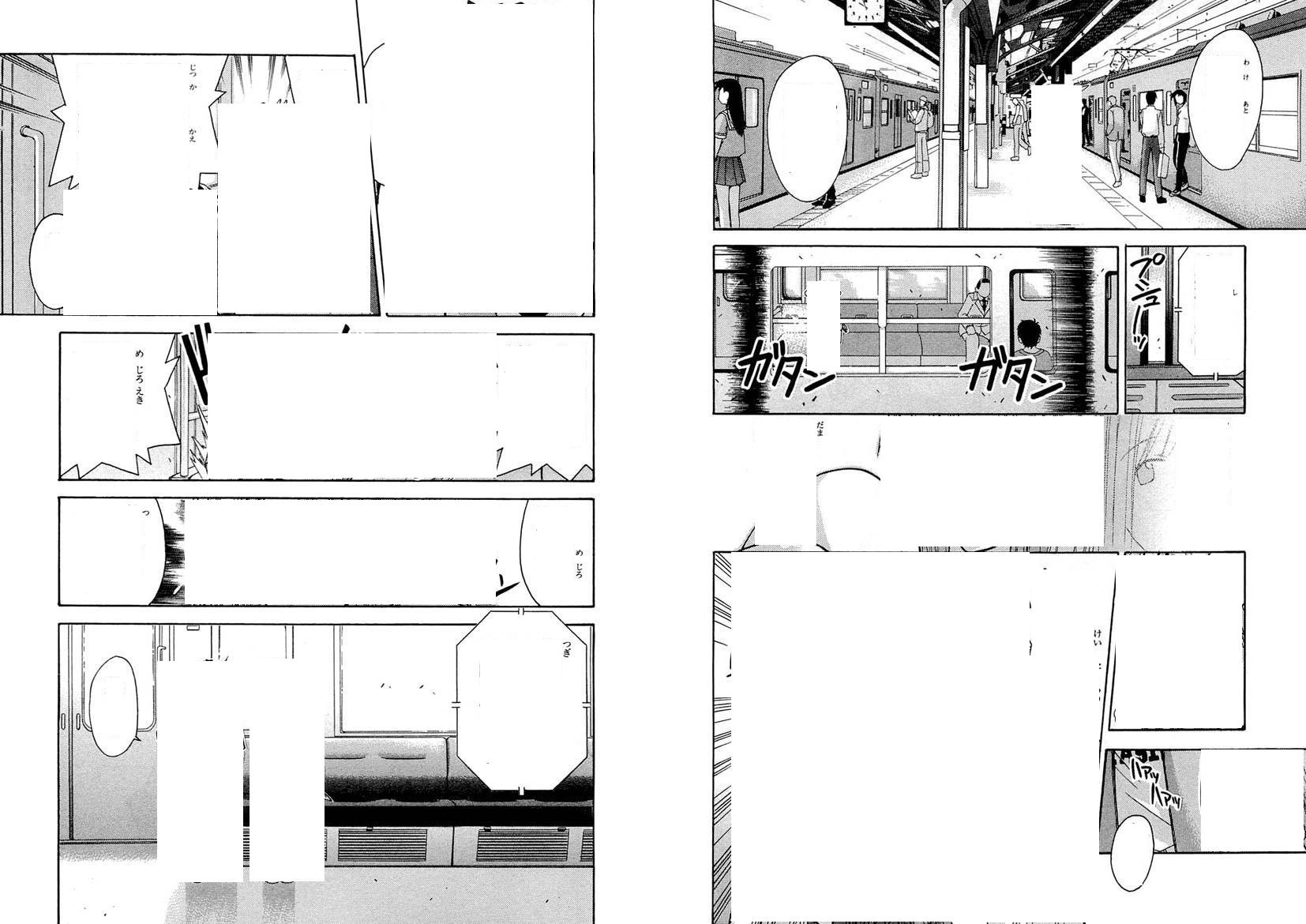}
        \caption{Masked}
        \label{fig:b}
    \end{subfigure}
    \hfill
    \begin{subfigure}[t]{0.15\textwidth}
        \centering
        \includegraphics[width=\textwidth]{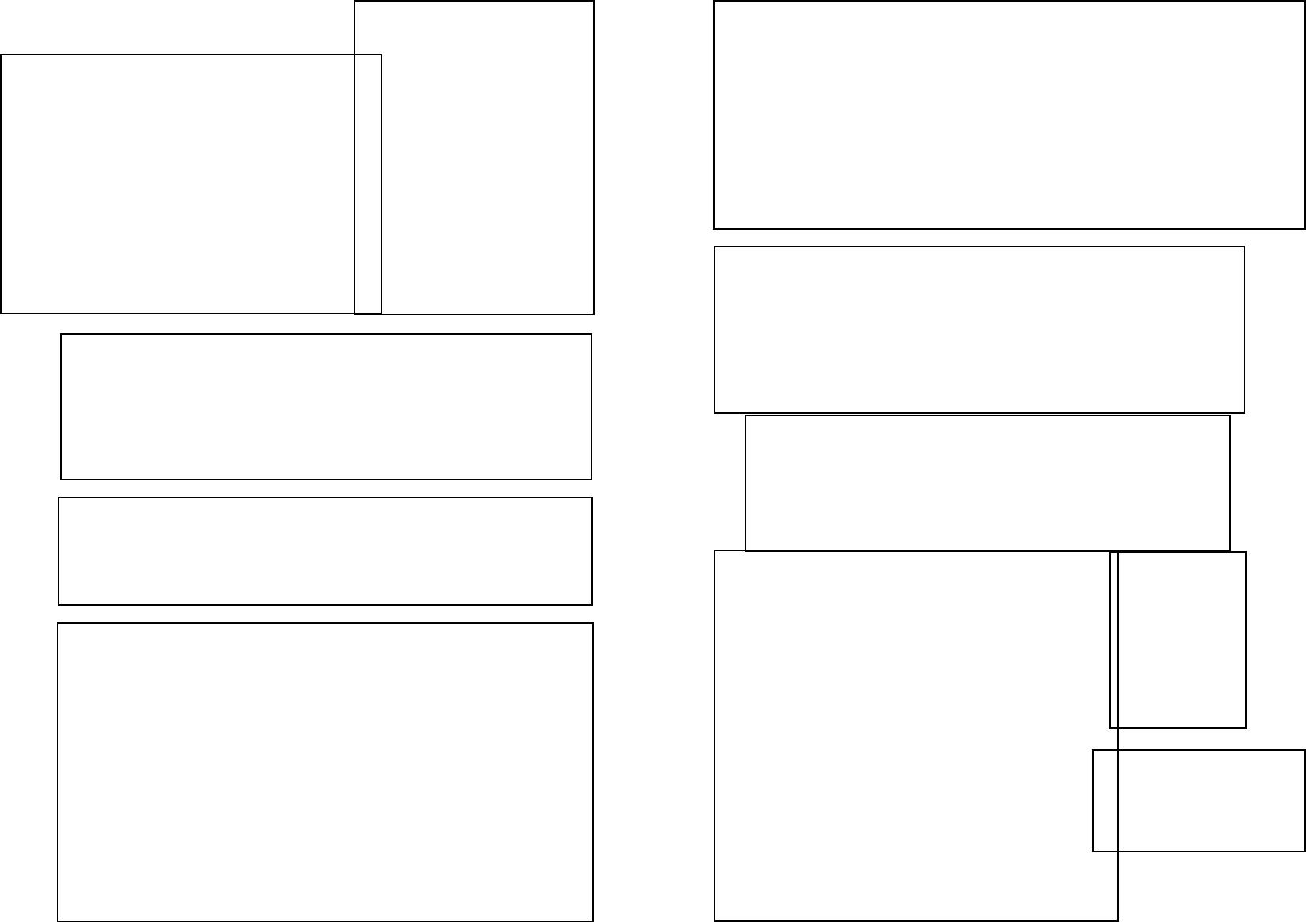}
        \caption{Panel frame-only}
        \label{fig:c}
    \end{subfigure}
    \caption{Examples of images used in the experiment. The images shown in the figure are from LoveHina \copyright Ken Akamatsu.}
    \label{fig:abcd}
\end{figure}

\begin{table}[t]
\centering
\caption{Overview of the classification dataset consisting of 104 works constructed from Manga109: * indicates that different volumes of the same work are included, resulting in a total of 104 works.}
\label{tab:datstats}
\begin{tabular}{l|cc}
\textbf{Genre} & \textbf{\makecell{Number of \\ works}} & \textbf{\makecell{Number of \\ pages with frames}} \\\hline
4-panel & 5 & 274 \\
Animal  & 5  & 479 \\
Battle & 9 & 823 \\
Fantasy & 12 & 1,154 \\
History & 6 & 699 \\
Horror & 2 & 178 \\
Humor & *15 & 1,391 \\
Love & *13 & 1,198 \\
Romantic comedy & *13 & 1,206 \\
SF & 14 & 1,294 \\
Sport & *10 & 968 \\
Suspense & 5 & 458 \\\hline
Total & *109 & 10,122
\end{tabular}
\end{table}

\begin{figure*}[t]
    \centering
    \begin{subfigure}[t]{0.32\textwidth} 
        \centering
        \includegraphics[width=\textwidth]{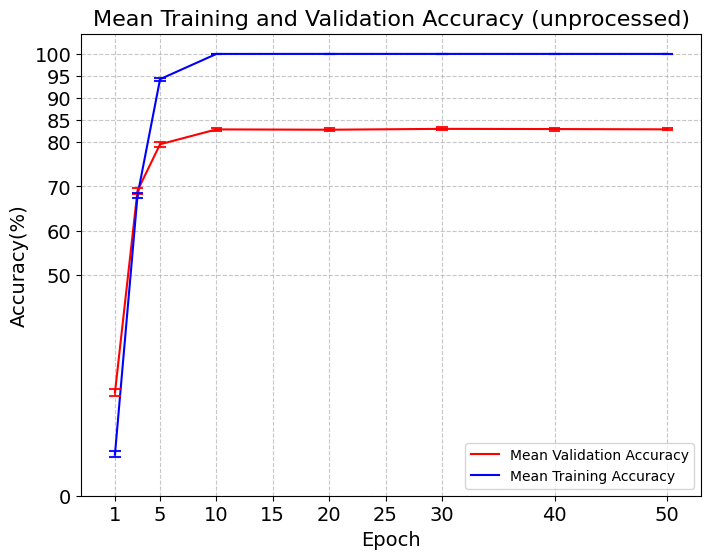}
        \caption*{Unprocessed} 
    \end{subfigure}
    \hspace{0.01\textwidth} 
    \begin{subfigure}[t]{0.32\textwidth} 
        \centering
        \includegraphics[width=\textwidth]{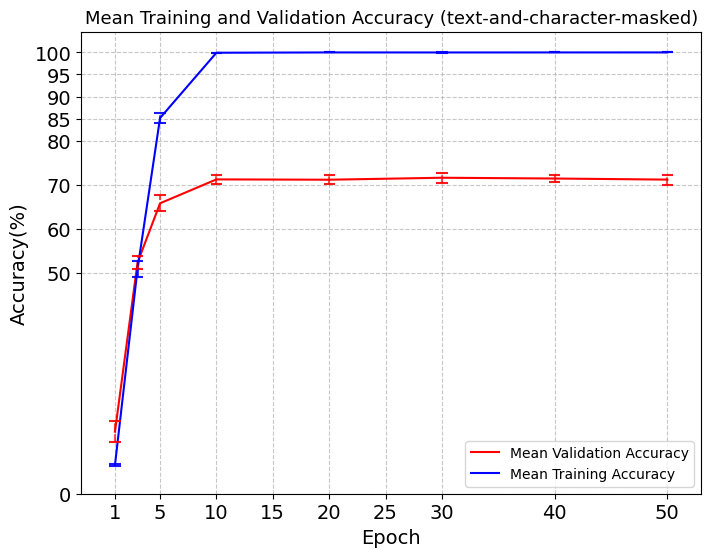} 
        \caption*{Text-and-character-masked}
    \end{subfigure}
    \hspace{0.01\textwidth} 
    \begin{subfigure}[t]{0.32\textwidth}
        \centering
        \includegraphics[width=\textwidth]{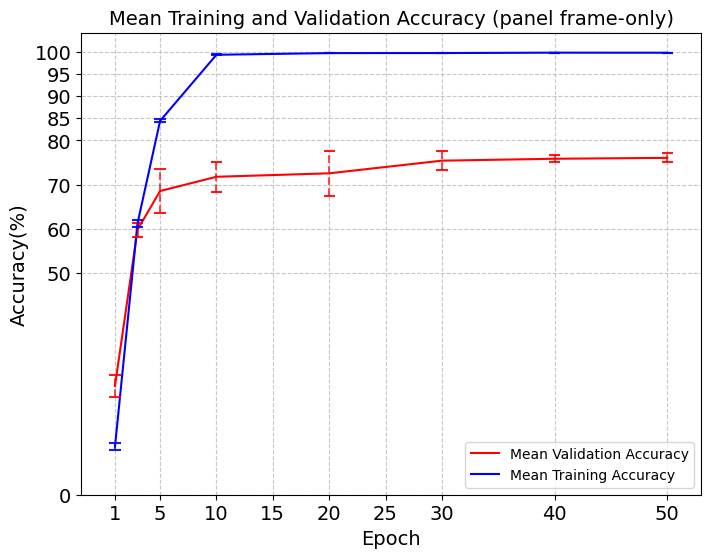} 
        \caption*{Panel frame-only}
    \end{subfigure}
    \caption{Mean training and validation accuracy curves during model training, with standard deviation represented by error bars.}
    \label{fig:accuracy}
\end{figure*}

In the experiments, we prepared unprocessed facing page images, masked facing page images, and facing page images with only the panel frames drawn. Figure~\ref{fig:a} shows the original unprocessed images. Using annotation data from Manga109, we applied masking to both characters and text, as shown in Figure~\ref{fig:b}. Additionally, we processed the images to retain only the panel frames, as shown in Figure~\ref{fig:c}. By using panel frame-only facing page images, we aimed to evaluate how strongly the panel layout features of each work contribute to the classification task.

\subsection{Multi-class Classification for 104 Works}

For data splitting, all page images were divided into a training set (80\%), development set (10\%), and test set (10\%), resulting in 8,053, 1,011, and 1,058 images, respectively. To enhance generalization, 5-fold cross-validation was applied to the training set, with one part used as validation and the rest for training. The final model performance was evaluated on the test set after fine-tuning parameters using the development set.

We utilized ResNet101, a convolutional neural network (CNN)~\citep{resnet, he2016identity} pretrained on ImageNet~\citep{imagenet2015}. Training used a mini-batch size of 32, cross-entropy loss, and Stochastic Gradient Descent (SGD) with a momentum factor of 90\%. The initial learning rate was 0.001, reduced by 90\% every 30 epochs. Training ran for up to 100 epochs, with classification experiments performed every epoch for the first 30 and every 10 epochs thereafter.

Figure~\ref{fig:accuracy} illustrates the classification accuracy of three models using different input types during the training process. Since the curves showed no significant changes after 50 epochs, only the first 50 epochs are displayed. This task is highly challenging, with a baseline classification accuracy of approximately 0.96\% when randomly selecting one class from 104 categories.

Testing on the test set was conducted using an ensemble voting approach with five models trained through 5-fold cross-validation. The final classification accuracies were 87.5\% for unprocessed images, 79.7\% for text-and-character-masked images, and 84.3\% for panel frame-only images. Notably, the ResNet101 model, pretrained on the ImageNet dataset of natural images, achieved relatively high classification accuracy on manga data. This suggests that manga’s visual features, particularly panel layouts, are sufficiently distinctive to be captured and leveraged by deep learning models, even those trained on vastly different domains.

Interestingly, the panel frame-only images yielded a relatively high accuracy of 84.3\%, even though they lack explicit visual elements such as characters and text. This demonstrates that panel layouts themselves contain inherent distinguishing characteristics unique to each manga work. These findings quantitatively confirm that spatial relationships, such as panel size, spacing, and alignment, play a critical role in distinguishing manga works.

However, classification accuracy does not decrease consistently with increasing restrictions on input features. For instance, the higher accuracy of panel frame-only images compared to text-and-character-masked images suggests that the masking process inadvertently obscures some panel layout information. As shown in Figure~\ref{fig:abcd}, the Manga109 dataset annotates text and character locations as rectangles. Characters often appear near panel edges, and masking these characters frequently obscures panel lines. This phenomenon highlights the importance of preserving complete panel information for accurate classification, further validating the strong distinguishing characteristics of panel layout designs in manga.

\subsection{Qualitative Analysis of Classification Results}

To better understand the classification process, we conducted a qualitative analysis using Grad-CAM~\citep{grad-cam}, a visualization technique that highlights image regions important for the model’s decision-making as heatmaps. This method reveals how the model interprets input types and identifies key features for classification.

\begin{figure}[t]
\centering
\begin{tabular}{c@{\hspace{0.3cm}}c} 
\includegraphics[width=0.2\textwidth]{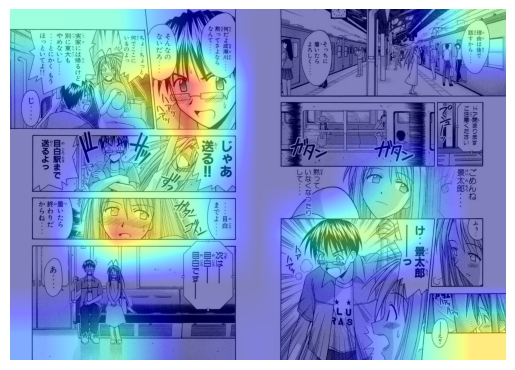} &
\includegraphics[width=0.2\textwidth]{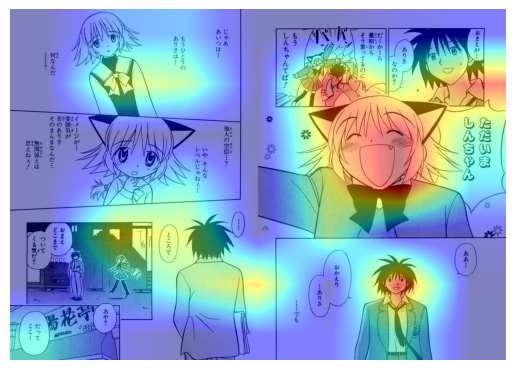} \\
Page 049 of LoveHina & Page 016 of Arisa \\
\copyright Ken Akamatsu & \copyright Ken Yagami 
\end{tabular}
\caption{Grad-CAM heatmaps with unprocessed images. The model correctly classified input images and exhibited strong feature focus on areas where characters were drawn.}
\label{fig:original_vis}
\end{figure}

\begin{figure}[t]
\centering
\begin{tabular}{c@{\hspace{0.3cm}}c}
\includegraphics[width=0.2\textwidth]{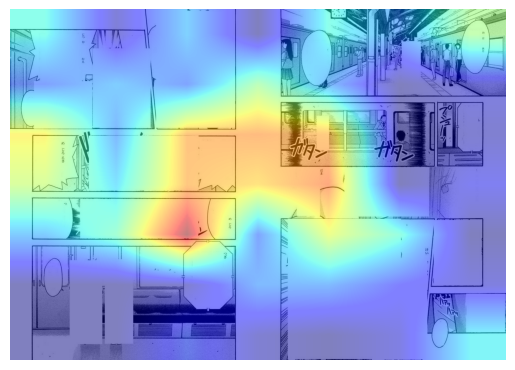} 
&
\includegraphics[width=0.2\textwidth]{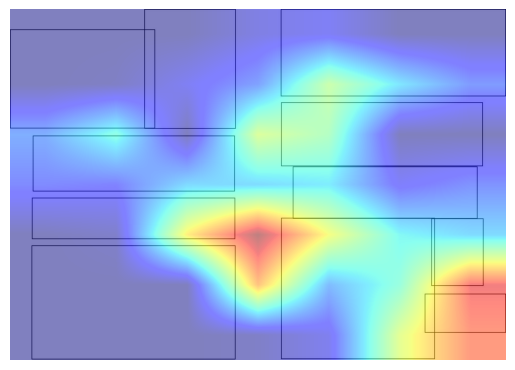}\\
Text-and-character-masked & Panel frame-only  \\
\end{tabular}
\caption{Grad-CAM heatmaps for page 49 of LoveHina\_vol14 \copyright Ken Akamatsu, with text-and-character-masked facing page images and panel frame-only facing page images.}
\label{fig:frame_only&masked_vis}
\end{figure}

Figure~\ref{fig:original_vis} shows Grad-CAM heatmaps for unprocessed images of two correctly classified manga works, where the model focuses primarily on characters, reaffirming their importance as distinguishing features in manga. Figure~\ref{fig:frame_only&masked_vis} highlights heatmaps for page 49 of LoveHina\_vol14, focusing on text-and-character-masked images and panel frame-only images. For masked images, the model shifts attention to panel layout details, such as border spacing and alignment, adapting to the absence of characters. In panel frame-only images, the model highlights two small, distinct frames at the bottom-right corner, commonly seen in LoveHina.

However, in text-and-character-masked images, these unique frames are frequently obscured due to the masking process in the Manga109 dataset, which uses rectangular annotations for text and character locations. As shown in Figure~\ref{fig:frame_only&masked_vis}, the masking unintentionally removes the borders of these smaller frames, preventing the model from utilizing this distinctive layout information. This limitation provides a plausible explanation for the lower classification accuracy of text-and-character-masked images compared to panel frame-only images.

These observations underline the importance of panel layouts as a unique visual feature. The model’s ability to accurately classify panel frame-only images demonstrates the strong discriminative power of layout-specific features, such as panel size, spacing, and alignment. 

Building on this understanding, we further investigated whether the subtle differences in panel layouts could reflect an author’s unique creative style, even within genres where such differences are difficult for human readers to discern.

To demonstrate that the panel layouts of each manga possess unique characteristics, we classified all five 4-panel manga works in the Manga109 dataset. The average classification accuracy was 93.1\%, with only one misclassification within the 4-panel manga category. This high accuracy suggests that even subtle differences in panel layouts, which are challenging for human judgment, can be effectively captured by the model. These differences may reflect the unique preferences of authors in arranging panel content, such as the amount of space allocated for characters or dialogue.

\begin{figure}[t]
\centering
\begin{tabular}{cc}
\includegraphics[width=0.2\textwidth]{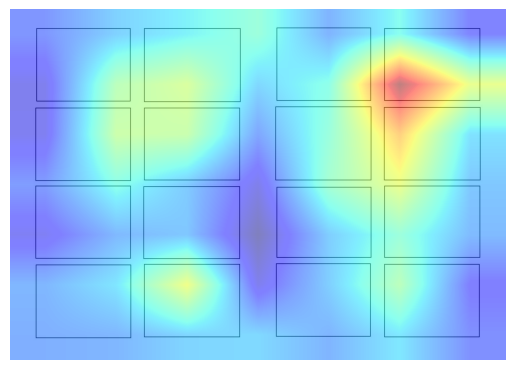}
&
\includegraphics[width=0.2\textwidth]{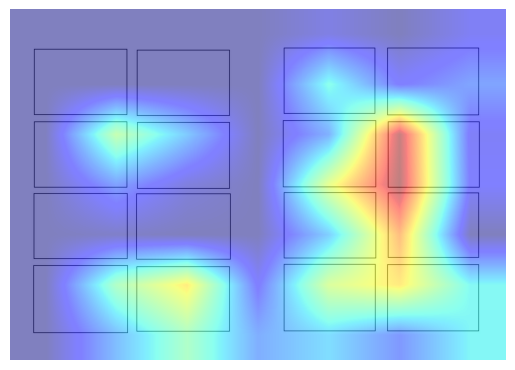} \\
Akuhamu & KoukouNoHitotachi \\
\copyright Satoshi Arai & \copyright Ani Kuzuhara
\end{tabular}
\caption{Grad-CAM heatmaps of the input images for the classification model (using panel frame-only facing page images). Both input images belong to 4-panel manga works, and the model classified them correctly.}
\label{fig:frame_vis}
\end{figure}

Next, the Grad-CAM heatmaps for classifying images from two 4-panel manga works are shown in Figure~\ref{fig:frame_vis}. The visualizations indicate that the model focuses on subtle layout features such as spacing between panels and the distribution of panels across the page. These features, while difficult for the human eye to distinguish, may be influenced by the author's preference for how much content (e.g. characters or speech balloons) they wish to include in each panel.

In conclusion, for panel frame-only facing page images, the Grad-CAM visualizations show that the model focuses solely on the spatial relationships and configurations of the panels. In the absence of any content, such as characters or text, the model relies entirely on panel layouts as distinguishing features for classification.

For 4-panel manga, which is particularly challenging for human judgment, the high classification accuracy achieved by the model highlights the subtle but critical differences in panel layouts. These differences, such as panel size, aspect ratio, spacing, and alignment, may reflect each author's unique approach to arranging content, such as characters or dialogue, within the panels. These layout decisions are not merely technical but also serve as expressions of the author's creative intent and storytelling style.

The high accuracy achieved across all manga works using panel frame-only facing page images supports the hypothesis that panel layouts possess inherent uniqueness across different manga works. Furthermore, it demonstrates that panel layouts play a critical role in distinguishing manga, even within genres like 4-panel manga, where differences are not easily perceived by the human eye. This finding underscores the strong discriminative power of layout features in manga classification tasks and their potential as a window into understanding the creative diversity of manga authors.

\subsection{Publisher Classification and Genre Classification}

In previous experiments, we observed that models focus on features like the spacing between panel frames and their distance to page edges. To determine whether these subtle features reflect the unique style of individual manga works, are influenced by publishers, or dictated by manga genres, we trained separate CNN models to classify manga by publisher and by genre.

From the Manga109 dataset, we selected 12 publishers (excluding two that contributed only one manga each). For the publisher classification task, one manga from each publisher was randomly chosen as the test set, with the remaining works as the training set. This setup provided sufficient training data while preventing the model from learning stylistic features specific to the test set. We used panel frame-only facing page images, with training methods and parameters identical to previous experiments.

As shown in Table~\ref{tab:publisher}, the publisher classification experiment achieved an average accuracy of 12.6\%, compared to 8.3\% for random guessing. These results suggest that publishers have minimal influence on panel layout design, indicating that panel layouts are not dictated by editorial policies but are instead unique to each manga work.

\begin{table}[t]
\centering
\caption{The results of the publisher classification experiment.}
\label{tab:publisher}
\begin{tabular}{l|ccc}
\textbf{Publisher} & \textbf{Precision} & \textbf{Recall} & \textbf{F1-score} \\\hline
ENIX  & 0.00 & 0.00 & 0.00\\
Hakusensha  & 0.12  & 0.07 & 0.09\\
Asahi Sonorama & 1.00 & 0.07 & 0.13\\
Tokuma Shoten & 0.34 & 0.37 & 0.35\\
TokyoSanseisya & 0.09 & 0.42 & 0.15\\
Shueisha & 0.14 & 0.29 & 0.19\\
Kodansha & 0.07 & 0.27 & 0.12\\
Kadokawa Shoten & 0.09 & 0.02 & 0.03\\
Akita Publishing & 0.05 & 0.01 & 0.02\\
Shogakukan & 0.15 & 0.12 & 0.13\\
Gakken Educational & 0.00 & 0.00 & 0.00\\
TAKESHOBO & 0.00 & 0.00 & 0.00\\%\hline
\end{tabular}
\end{table}

In Table~\ref{tab:datstats}, we have already seen the distribution of different manga genres in the Manga109 dataset. The dataset was split in the same manner as in the publisher classification experiment, ensuring that the model did not learn unique features specific to the test set during training. Training methods and parameters were again consistent with those used in previous experiments.

The results of the genre classification experiment are presented in Table~\ref{tab:genre}. The overall average classification accuracy was 20.8\%. Notably, the model demonstrated a relatively higher classification ability for 4-panel manga, which adheres to strict panel layout rules, as compared to other genres. For some genres, such as Humor and Love, the model showed moderate classification ability, likely due to the long histories of these genres, which may have led to the development of more established panel layout conventions. However, for other genres, the model struggled to classify the works effectively. When compared to the significantly higher average classification accuracy of 84.3\% achieved in predicting specific manga titles based on panel layouts, these results indicate that manga genres contribute relatively less to panel layout features. This reinforces the conclusion that panel layouts are distinct and unique characteristics of individual manga works.

\begin{table}[t]
\centering
\caption{The results of the genre classification experiment.}
\label{tab:genre}
\begin{tabular}{l|ccc}
\textbf{Genre} & \textbf{Precision} & \textbf{Recall} & \textbf{F1-score} \\\hline
4-panel & 0.89 & 0.70 & 0.78 \\
Animal  & 0.00  & 0.00 & 0.00\\
Battle & 0.02 & 0.01 & 0.01\\
Fantasy & 0.00 & 0.00 & 0.00\\
History & 0.06 & 0.01 & 0.02\\
Horror & 0.00 & 0.00 & 0.00\\
Humor & 0.34 & 0.67 & 0.46\\
Love & 0.38 & 0.74 & 0.50\\
Romantic comedy & 0.01 & 0.03 & 0.02\\
SF & 0.30 & 0.49 & 0.38\\
Sport & 0.01 & 0.02 & 0.02\\
Suspense & 0.00 & 0.00 & 0.00\\%\hline
\end{tabular}
\end{table}

In conclusion, the publisher and genre classification experiments revealed that neither publishers nor genres significantly influence the panel layout of manga. The details of panel layouts, such as panel size, spacing between borders, and the spacing between borders and page edges, are likely determined by the creative choices of individual manga authors. These findings further emphasize the individuality of panel layouts as a defining feature of manga works.

\subsection{Classification Experiments with Noisy Panel Frame-only Facing Page Images}

In the previous section, we confirmed that panel layout features are unique to each manga and are not determined by the publisher or genre. To further investigate the role of panel layout features in manga classification, we conducted additional experiments using noisy panel frame-only facing page images as input. In these experiments, random noise was introduced to disrupt the spatial relationships between panels while retaining the overall layout structure to some extent. This approach allowed us to test whether the classification accuracy would significantly decrease, thereby validating the importance of precise panel layout information.
We employed two types of noise, as illustrated in Figure~\ref{fig:noise}:

\subsubsection{Rectangular noisy panels (Figure~\ref{fig:noise_rectangle}):}
\begin{figure}[htbp]
    \centering
    \begin{subfigure}[t]{0.2\textwidth} 
        \centering
        \includegraphics[width=\textwidth]{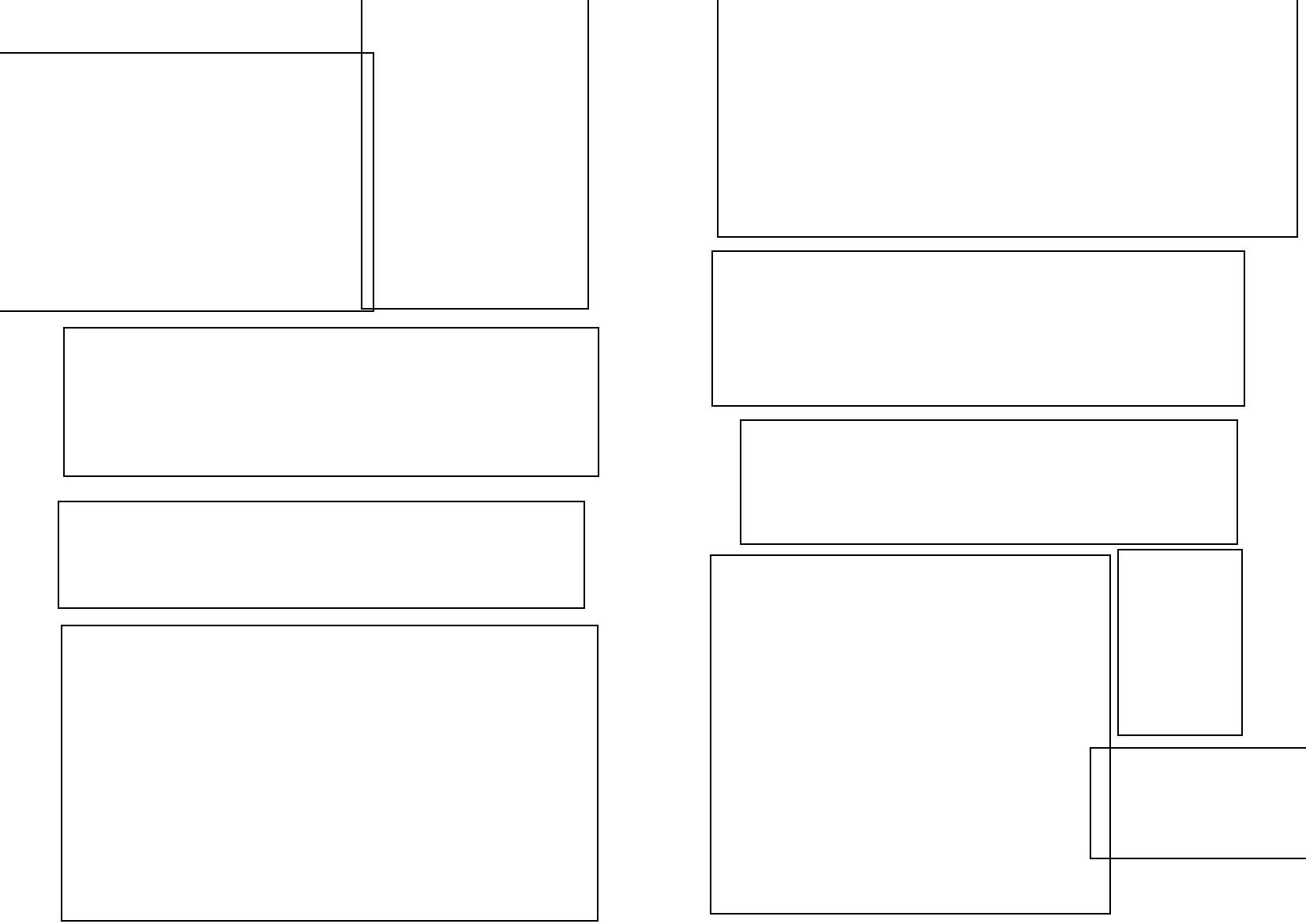} 
        \caption{Rectangle noisy panels.} 
        \label{fig:noise_rectangle}
    \end{subfigure}
    \hfill
    \begin{subfigure}[t]{0.2\textwidth}
        \centering
        \includegraphics[width=\textwidth]{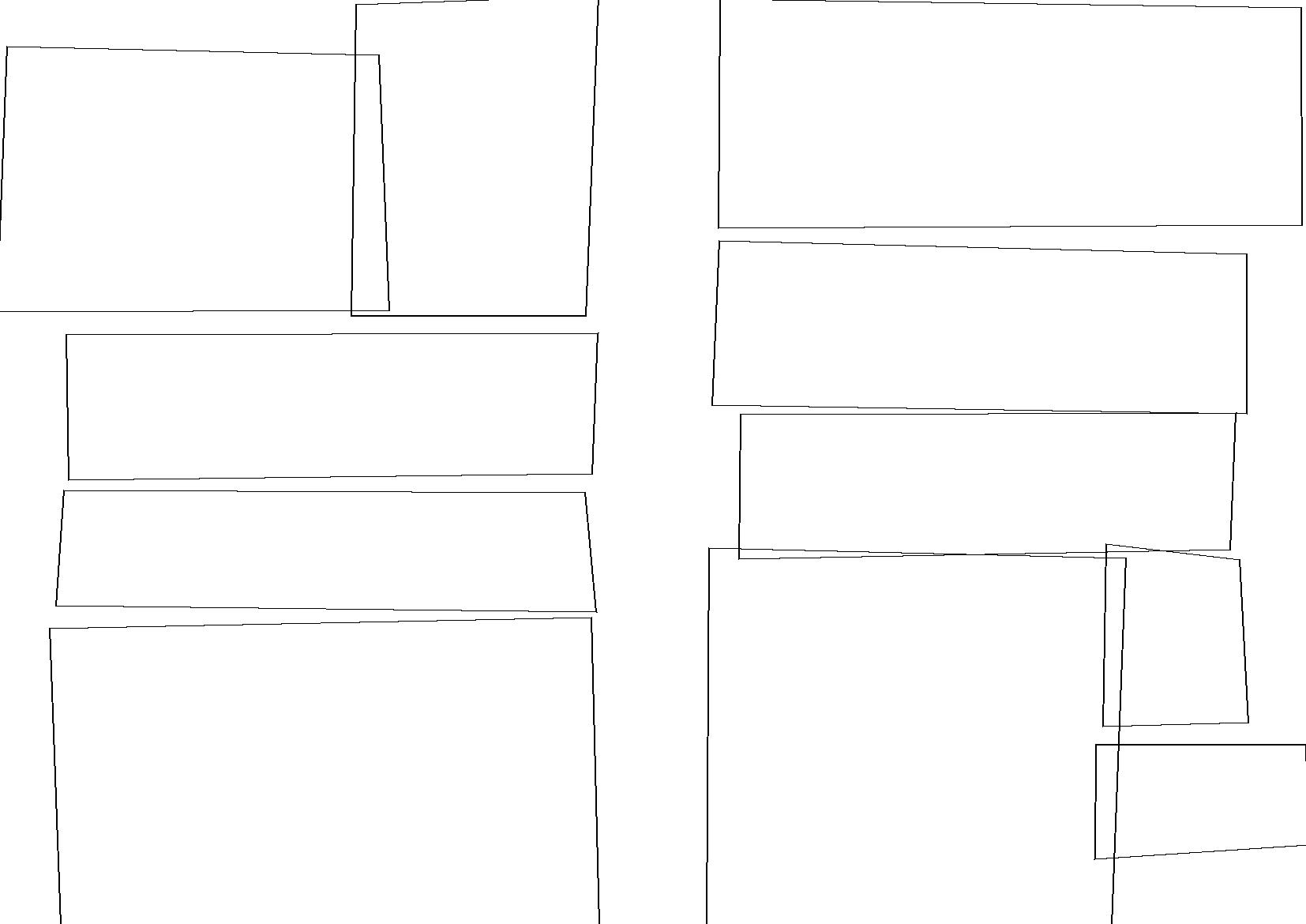}
        \caption{Quadrilateral noisy panels.}
        \label{fig:noise_quadrilateral}
    \end{subfigure}
    \caption{Noisy image examples.}
    \label{fig:noise}
\end{figure}

The first type retained the rectangular characteristics of the panels but randomly shifted their vertices by a certain value. Two experiments were conducted by adding noise of (-10, 10) and (-20, 20) to the vertices based on their original positions.
\subsubsection{Quadrilateral noisy panels (Figure~\ref{fig:noise_quadrilateral}): }
The second type only retained the quadrilateral characteristics of the panels, with vertices randomly shifted by a certain value. Similarly, two experiments were conducted with noise of (-10, 10) and (-20, 20).

\begin{table}[ht]
\centering
\caption{Classification accuracy on noisy panel frame-only images.}
\label{tab:noisy}
\begin{tabular}{l|rr}

\textbf{Noise Type} & \textbf{Noise Range} & \textbf{Accuracy} \\
\hline
Rectangular         & (-10, 10)            & 0.67 \\
Rectangular         & (-20, 20)            & 0.53 \\
Quadrilateral       & (-10, 10)            & 0.71 \\
Quadrilateral       & (-20, 20)            & 0.59 \\

\end{tabular}
\end{table}

We further added noise to panel frame positions to test the model's sensitivity to layout distortion. As shown in Table~\ref{tab:noisy}, the classification accuracy decreased as the noise level increased, indicating that precise panel positioning plays a meaningful role in layout-based classification. Although classification accuracy for noisy panel frame images was lower than that for original panel frame images, the model still demonstrated a certain level of accuracy. These results indicate that while noise reduces the model's ability to utilize panel layout features, the distinctive characteristics of panel designs across works remain sufficient for classification, even under noisy conditions.

Interestingly, the classification accuracy for quadrilateral noisy panels was consistently higher than for rectangular noisy panels. We hypothesize that this is because adding noise to rectangular panels disrupts unique characteristics of the original work more significantly, such as the spacing between panels and the margins between panels and page edges. In contrast, quadrilateral noisy panels preserve more of the original layout features, supporting higher classification accuracy.

\begin{figure}[htbp]
    \centering
    \begin{subfigure}[t]{0.15\textwidth}
        \centering
        \includegraphics[width=\textwidth]{figure/049_LoveHina_vol14_frame_only_heatmap4.0.png} 
        \caption{Unprocessed}
        \label{fig:unprocessed frame}
    \end{subfigure}
    \hfill
    \begin{subfigure}[t]{0.15\textwidth}
        \centering
        \includegraphics[width=\textwidth]{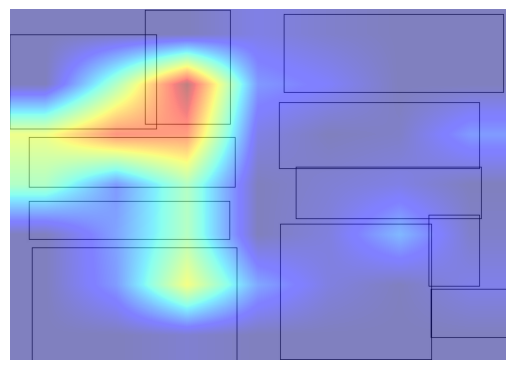}
        \caption{Rectangle noise}
        \label{fig:rectangle}
    \end{subfigure}
    \hfill
    \begin{subfigure}[t]{0.15\textwidth}
        \centering
        \includegraphics[width=\textwidth]{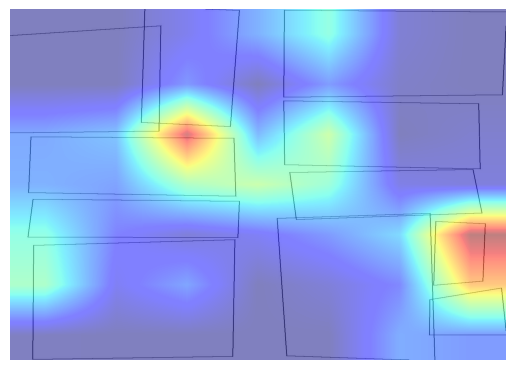}
        \caption{Quadrilateral noise}
        \label{fig:quadrilateral}
    \end{subfigure}
    \caption{Three types of panel frame-only facing page image.}
    \label{fig:noise-CAM}
\end{figure}

For instance, consider Figure~\ref{fig:noise-CAM}, which shows three panel frame-only images based on the original page 49 of LoveHina\_vol14, with Grad-CAM heatmaps overlaid. The model trained on the rectangle noisy panel frame image failed to classify correctly. In both Figure~\ref{fig:unprocessed frame} and Figure~\ref{fig:quadrilateral}, the model strongly focuses on the spacing between the small bottom-right panel and the page edge. This slightly outward-protruding bottom-right corner panel, relative to the row above, appears to be a distinctive layout feature of LoveHina. When noise is added while retaining rectangular characteristics, such layout features are more likely to be lost compared to when only quadrilateral characteristics are preserved.

These findings provide further evidence that panel layouts in manga, even in subtle details that are challenging for human perception, play a decisive role in distinguishing works. This experiment highlights that the spatial relationships and configurations of panels are deeply tied to the unique characteristics of each manga, reinforcing the importance of panel layouts as a defining feature in manga classification tasks.

\section{Conclusion}
This study used deep learning to explore whether panel page designs in manga vary by work. Our experiments showed that even without characters and text, panel layouts exhibit work-specific uniqueness, enabling reliable classification and highlighting their role as a distinguishing visual structure in manga. This was validated through classification tasks and supported by Grad-CAM visualizations.

Additional experiments confirmed that panel layouts are minimally influenced by publishers or genres, underscoring their individuality at the work level. Even with added noise, panel layouts retained sufficient distinctiveness for classification, emphasizing their intricate role in manga creation.

While our study isolates panel layouts, we acknowledge that other visual elements—such as character design, gaze direction, and speech balloon positioning—also influence manga style and storytelling. Future research could investigate the interplay between these features and layout.

These findings highlight the significance of panel layouts as a unique visual feature in manga, reflecting the stylistic diversity of individual works and their contribution to storytelling structure.
Future research directions include conducting experiments with a larger dataset to verify whether classification accuracy can remain high as the number of categories increases. Additionally, due to current data limitations, the differences in page design between works by the same author have not been thoroughly explored. Further investigation into this question is planned for future studies.

\section*{Acknowledgments}
This work was supported by JSPS KAKENHI Grant Number JP24K02993.

\bibliographystyle{apacite} % 或其他样式，如 unsrtnat 或 abbrvnat
\bibliography{CogSci_Template} % 这里填入您的 .bib 文件名称（不加扩展名）

\begin{thebibliography}{}

\bibitem [\protect \citeauthoryear {%
Augereau%
, Iwata%
\BCBL {}\ \BBA {} Kise%
}{%
Augereau%
\ \protect \BOthers {.}}{%
{\protect \APACyear {2018}}%
}]{%
augereau2018survey}
\APACinsertmetastar {%
augereau2018survey}%
\begin{APACrefauthors}%
Augereau, O.%
, Iwata, M.%
\BCBL {}\ \BBA {} Kise, K.%
\end{APACrefauthors}%
\unskip\
\newblock
\APACrefYearMonthDay{2018}{}{}.
\newblock
{\BBOQ}\APACrefatitle {A survey of comics research in computer science} {A survey of comics research in computer science}.{\BBCQ}
\newblock
\APACjournalVolNumPages{Journal of imaging}{4}{7}{87}.
\PrintBackRefs{\CurrentBib}

\bibitem [\protect \citeauthoryear {%
Chu%
\ \BBA {} Guo%
}{%
Chu%
\ \BBA {} Guo%
}{%
{\protect \APACyear {2017}}%
}]{%
chu-guo}
\APACinsertmetastar {%
chu-guo}%
\begin{APACrefauthors}%
Chu, W\BHBI T.%
\BCBT {}\ \BBA {} Guo, H\BHBI J.%
\end{APACrefauthors}%
\unskip\
\newblock
\APACrefYearMonthDay{2017}{}{}.
\newblock
{\BBOQ}\APACrefatitle {Movie Genre Classification Based on Poster Images with Deep Neural Networks} {Movie genre classification based on poster images with deep neural networks}.{\BBCQ}
\newblock
\BIn{} \APACrefbtitle {Proceedings of the Workshop on Multimodal Understanding of Social, Affective and Subjective Attributes} {Proceedings of the workshop on multimodal understanding of social, affective and subjective attributes}\ (\BPG~39-45).
\newblock
\APACaddressPublisher{}{Association for Computing Machinery}.
\PrintBackRefs{\CurrentBib}

\bibitem [\protect \citeauthoryear {%
Cohn%
}{%
Cohn%
}{%
{\protect \APACyear {2013}}%
}]{%
cohn2013visual}
\APACinsertmetastar {%
cohn2013visual}%
\begin{APACrefauthors}%
Cohn, N.%
\end{APACrefauthors}%
\unskip\
\newblock
\APACrefYear{2013}.
\newblock
\APACrefbtitle {The Visual Language of Comics: Introduction to the Structure and Cognition of Sequential Images} {The visual language of comics: Introduction to the structure and cognition of sequential images}.
\newblock
\APACaddressPublisher{}{Bloomsbury}.
\PrintBackRefs{\CurrentBib}

\bibitem [\protect \citeauthoryear {%
Dubray%
\ \BBA {} Laubrock%
}{%
Dubray%
\ \BBA {} Laubrock%
}{%
{\protect \APACyear {2019}}%
}]{%
balloon}
\APACinsertmetastar {%
balloon}%
\begin{APACrefauthors}%
Dubray, D.%
\BCBT {}\ \BBA {} Laubrock, J.%
\end{APACrefauthors}%
\unskip\
\newblock
\APACrefYearMonthDay{2019}{}{}.
\newblock
{\BBOQ}\APACrefatitle {Deep CNN-based Speech Balloon Detection and Segmentation for Comic Books} {Deep cnn-based speech balloon detection and segmentation for comic books}.{\BBCQ}
\newblock
\APACjournalVolNumPages{CoRR}{abs/1902.08137}{}{}.
\PrintBackRefs{\CurrentBib}

\bibitem [\protect \citeauthoryear {%
Dunst%
, Hartel%
\BCBL {}\ \BBA {} Laubrock%
}{%
Dunst%
\ \protect \BOthers {.}}{%
{\protect \APACyear {2017}}%
}]{%
gnc}
\APACinsertmetastar {%
gnc}%
\begin{APACrefauthors}%
Dunst, A.%
, Hartel, R.%
\BCBL {}\ \BBA {} Laubrock, J.%
\end{APACrefauthors}%
\unskip\
\newblock
\APACrefYearMonthDay{2017}{}{}.
\newblock
{\BBOQ}\APACrefatitle {The Graphic Narrative Corpus (GNC): Design, Annotation, and Analysis for the Digital Humanities} {The graphic narrative corpus (gnc): Design, annotation, and analysis for the digital humanities}.{\BBCQ}
\newblock
\BIn{} \APACrefbtitle {2017 14th IAPR International Conference on Document Analysis and Recognition (ICDAR)} {2017 14th iapr international conference on document analysis and recognition (icdar)}\ (\BVOL~03, \BPG~15-20).
\PrintBackRefs{\CurrentBib}

\bibitem [\protect \citeauthoryear {%
Fujimoto%
\ \protect \BOthers {.}}{%
Fujimoto%
\ \protect \BOthers {.}}{%
{\protect \APACyear {2016}}%
}]{%
manga109}
\APACinsertmetastar {%
manga109}%
\begin{APACrefauthors}%
Fujimoto, A.%
, Ogawa, T.%
, Yamamoto, K.%
, Matsui, Y.%
, Yamasaki, T.%
\BCBL {}\ \BBA {} Aizawa, K.%
\end{APACrefauthors}%
\unskip\
\newblock
\APACrefYearMonthDay{2016}{}{}.
\newblock
{\BBOQ}\APACrefatitle {Manga109 dataset and creation of metadata} {Manga109 dataset and creation of metadata}.{\BBCQ}
\newblock
\BIn{} \APACrefbtitle {Proceedings of the 1st international workshop on comics analysis, processing and understanding} {Proceedings of the 1st international workshop on comics analysis, processing and understanding}\ (\BPGS\ 1--5).
\PrintBackRefs{\CurrentBib}

\bibitem [\protect \citeauthoryear {%
Godwin%
, Eng%
, Todaro%
, Murray%
\BCBL {}\ \BBA {} Fisher%
}{%
Godwin%
\ \protect \BOthers {.}}{%
{\protect \APACyear {2018}}%
}]{%
godwin2018examination}
\APACinsertmetastar {%
godwin2018examination}%
\begin{APACrefauthors}%
Godwin, K\BPBI E.%
, Eng, C\BPBI M.%
, Todaro, R.%
, Murray, G.%
\BCBL {}\ \BBA {} Fisher, A\BPBI V.%
\end{APACrefauthors}%
\unskip\
\newblock
\APACrefYearMonthDay{2018}{}{}.
\newblock
{\BBOQ}\APACrefatitle {Examination of the Role of Book Layout, Executive Function, and Processing Speed On Children's Decoding and Reading Comprehension.} {Examination of the role of book layout, executive function, and processing speed on children's decoding and reading comprehension.}{\BBCQ}
\newblock
\BIn{} \APACrefbtitle {Proceedings of the Annual Meeting of the Cognitive Science Society} {Proceedings of the annual meeting of the cognitive science society}\ (\BVOL~40).
\PrintBackRefs{\CurrentBib}

\bibitem [\protect \citeauthoryear {%
He%
, Zhang%
, Ren%
\BCBL {}\ \BBA {} Sun%
}{%
He%
\ \protect \BOthers {.}}{%
{\protect \APACyear {2015}}%
}]{%
resnet}
\APACinsertmetastar {%
resnet}%
\begin{APACrefauthors}%
He, K.%
, Zhang, X.%
, Ren, S.%
\BCBL {}\ \BBA {} Sun, J.%
\end{APACrefauthors}%
\unskip\
\newblock
\APACrefYearMonthDay{2015}{}{}.
\newblock
{\BBOQ}\APACrefatitle {Deep Residual Learning for Image Recognition} {Deep residual learning for image recognition}.{\BBCQ}
\newblock
\APACjournalVolNumPages{CoRR}{abs/1512.03385}{}{}.
\PrintBackRefs{\CurrentBib}

\bibitem [\protect \citeauthoryear {%
He%
, Zhang%
, Ren%
\BCBL {}\ \BBA {} Sun%
}{%
He%
\ \protect \BOthers {.}}{%
{\protect \APACyear {2016}}%
}]{%
he2016identity}
\APACinsertmetastar {%
he2016identity}%
\begin{APACrefauthors}%
He, K.%
, Zhang, X.%
, Ren, S.%
\BCBL {}\ \BBA {} Sun, J.%
\end{APACrefauthors}%
\unskip\
\newblock
\APACrefYearMonthDay{2016}{}{}.
\newblock
{\BBOQ}\APACrefatitle {Identity Mappings in Deep Residual Networks} {Identity mappings in deep residual networks}.{\BBCQ}
\newblock
\BIn{} \APACrefbtitle {Proceedings of the European Conference on Computer Vision (ECCV)} {Proceedings of the european conference on computer vision (eccv)}\ (\BPGS\ 630--645).
\PrintBackRefs{\CurrentBib}

\bibitem [\protect \citeauthoryear {%
Kukreja%
\ \protect \BOthers {.}}{%
Kukreja%
\ \protect \BOthers {.}}{%
{\protect \APACyear {2023}}%
}]{%
kukreja2023unmasking}
\APACinsertmetastar {%
kukreja2023unmasking}%
\begin{APACrefauthors}%
Kukreja, V.%
\BCBT {}\ \BOthersPeriod {.}
\end{APACrefauthors}%
\unskip\
\newblock
\APACrefYearMonthDay{2023}{}{}.
\newblock
{\BBOQ}\APACrefatitle {Unmasking Comic Narratives: CNN-RF Approach to Genre Classification} {Unmasking comic narratives: Cnn-rf approach to genre classification}.{\BBCQ}
\newblock
\BIn{} \APACrefbtitle {2023 4th International Conference on Data Analytics for Business and Industry (ICDABI)} {2023 4th international conference on data analytics for business and industry (icdabi)}\ (\BPGS\ 555--560).
\PrintBackRefs{\CurrentBib}

\bibitem [\protect \citeauthoryear {%
Laubrock%
\ \BBA {} Dubray%
}{%
Laubrock%
\ \BBA {} Dubray%
}{%
{\protect \APACyear {2019}}%
}]{%
laubrock}
\APACinsertmetastar {%
laubrock}%
\begin{APACrefauthors}%
Laubrock, J.%
\BCBT {}\ \BBA {} Dubray, D.%
\end{APACrefauthors}%
\unskip\
\newblock
\APACrefYearMonthDay{2019}{01}{}.
\newblock
{\BBOQ}\APACrefatitle {CNN-Based Classification of Illustrator Style in Graphic Novels: Which Features Contribute Most?: 25th International Conference, MMM 2019, Thessaloniki, Greece, January 8–11, 2019, Proceedings, Part II} {Cnn-based classification of illustrator style in graphic novels: Which features contribute most?: 25th international conference, mmm 2019, thessaloniki, greece, january 8–11, 2019, proceedings, part ii}.{\BBCQ}
\newblock
\BIn{} (\BPG~684-695).
\PrintBackRefs{\CurrentBib}

\bibitem [\protect \citeauthoryear {%
Laubrock%
\ \BBA {} Dunst%
}{%
Laubrock%
\ \BBA {} Dunst%
}{%
{\protect \APACyear {2020}}%
}]{%
laubrock2020computational}
\APACinsertmetastar {%
laubrock2020computational}%
\begin{APACrefauthors}%
Laubrock, J.%
\BCBT {}\ \BBA {} Dunst, A.%
\end{APACrefauthors}%
\unskip\
\newblock
\APACrefYearMonthDay{2020}{}{}.
\newblock
{\BBOQ}\APACrefatitle {Computational approaches to comics analysis} {Computational approaches to comics analysis}.{\BBCQ}
\newblock
\APACjournalVolNumPages{Topics in cognitive science}{12}{1}{274--310}.
\PrintBackRefs{\CurrentBib}

\bibitem [\protect \citeauthoryear {%
Matsui%
\ \protect \BOthers {.}}{%
Matsui%
\ \protect \BOthers {.}}{%
{\protect \APACyear {2017}}%
}]{%
manga109-original}
\APACinsertmetastar {%
manga109-original}%
\begin{APACrefauthors}%
Matsui, Y.%
, Ito, K.%
, Aramaki, Y.%
, Fujimoto, A.%
, Ogawa, T.%
, Yamasaki, T.%
\BCBL {}\ \BBA {} Aizawa, K.%
\end{APACrefauthors}%
\unskip\
\newblock
\APACrefYearMonthDay{2017}{}{}.
\newblock
{\BBOQ}\APACrefatitle {Sketch-based Manga Retrieval using Manga109 Dataset} {Sketch-based manga retrieval using manga109 dataset}.{\BBCQ}
\newblock
\APACjournalVolNumPages{Multimedia Tools and Applications}{76}{20}{21811--21838}.
\PrintBackRefs{\CurrentBib}

\bibitem [\protect \citeauthoryear {%
McCloud%
\ \BBA {} Martin%
}{%
McCloud%
\ \BBA {} Martin%
}{%
{\protect \APACyear {1993}}%
}]{%
mccloud1993understanding}
\APACinsertmetastar {%
mccloud1993understanding}%
\begin{APACrefauthors}%
McCloud, S.%
\BCBT {}\ \BBA {} Martin, M.%
\end{APACrefauthors}%
\unskip\
\newblock
\APACrefYear{1993}.
\newblock
\APACrefbtitle {Understanding comics: The invisible art} {Understanding comics: The invisible art}\ (\BVOL~106).
\newblock
\APACaddressPublisher{}{Kitchen sink press Northampton, MA}.
\PrintBackRefs{\CurrentBib}

\bibitem [\protect \citeauthoryear {%
Natsume%
, Holt%
\BCBL {}\ \BBA {} Fukuda%
}{%
Natsume%
\ \protect \BOthers {.}}{%
{\protect \APACyear {2020}}%
}]{%
fusanosuke2020panel}
\APACinsertmetastar {%
fusanosuke2020panel}%
\begin{APACrefauthors}%
Natsume, F.%
, Holt, J.%
\BCBL {}\ \BBA {} Fukuda, T.%
\end{APACrefauthors}%
\unskip\
\newblock
\APACrefYearMonthDay{2020}{}{}.
\newblock
{\BBOQ}\APACrefatitle {Panel Configurations in Sh{\=o}jo Manga} {Panel configurations in sh{\=o}jo manga}.{\BBCQ}
\newblock
\APACjournalVolNumPages{US-Japan Women's Journal}{58}{58}{58--74}.
\PrintBackRefs{\CurrentBib}

\bibitem [\protect \citeauthoryear {%
Rigaud%
, Nguyen%
\BCBL {}\ \BBA {} Burie%
}{%
Rigaud%
\ \protect \BOthers {.}}{%
{\protect \APACyear {2021}}%
}]{%
Rigaud}
\APACinsertmetastar {%
Rigaud}%
\begin{APACrefauthors}%
Rigaud, C.%
, Nguyen, N\BHBI V.%
\BCBL {}\ \BBA {} Burie, J\BHBI C.%
\end{APACrefauthors}%
\unskip\
\newblock
\APACrefYearMonthDay{2021}{}{}.
\newblock
{\BBOQ}\APACrefatitle {Text Block Segmentation in Comic Speech Bubbles} {Text block segmentation in comic speech bubbles}.{\BBCQ}
\newblock
\BIn{} \APACrefbtitle {Pattern Recognition. ICPR International Workshops and Challenges: Virtual Event, January 10–15, 2021, Proceedings, Part VI} {Pattern recognition. icpr international workshops and challenges: Virtual event, january 10–15, 2021, proceedings, part vi}\ (\BPG~250-261).
\newblock
\APACaddressPublisher{}{Springer-Verlag}.
\PrintBackRefs{\CurrentBib}

\bibitem [\protect \citeauthoryear {%
Russakovsky%
\ \protect \BOthers {.}}{%
Russakovsky%
\ \protect \BOthers {.}}{%
{\protect \APACyear {2015}}%
}]{%
imagenet2015}
\APACinsertmetastar {%
imagenet2015}%
\begin{APACrefauthors}%
Russakovsky, O.%
, Deng, J.%
, Su, H.%
, Krause, J.%
, Satheesh, S.%
, Ma, S.%
\BDBL {}Fei-Fei, L.%
\end{APACrefauthors}%
\unskip\
\newblock
\APACrefYearMonthDay{2015}{}{}.
\newblock
{\BBOQ}\APACrefatitle {ImageNet Large Scale Visual Recognition Challenge} {Imagenet large scale visual recognition challenge}.{\BBCQ}
\newblock
\APACjournalVolNumPages{International Journal of Computer Vision (IJCV)}{115}{3}{211--252}.
\PrintBackRefs{\CurrentBib}

\bibitem [\protect \citeauthoryear {%
Selvaraju%
\ \protect \BOthers {.}}{%
Selvaraju%
\ \protect \BOthers {.}}{%
{\protect \APACyear {2019}}%
}]{%
grad-cam}
\APACinsertmetastar {%
grad-cam}%
\begin{APACrefauthors}%
Selvaraju, R\BPBI R.%
, Cogswell, M.%
, Das, A.%
, Vedantam, R.%
, Parikh, D.%
\BCBL {}\ \BBA {} Batra, D.%
\end{APACrefauthors}%
\unskip\
\newblock
\APACrefYearMonthDay{2019}{oct}{}.
\newblock
{\BBOQ}\APACrefatitle {Grad-{CAM}: Visual Explanations from Deep Networks via Gradient-Based Localization} {Grad-{CAM}: Visual explanations from deep networks via gradient-based localization}.{\BBCQ}
\newblock
\APACjournalVolNumPages{International Journal of Computer Vision}{128}{2}{336--359}.
\PrintBackRefs{\CurrentBib}

\bibitem [\protect \citeauthoryear {%
Sharma%
\ \BBA {} Kukreja%
}{%
Sharma%
\ \BBA {} Kukreja%
}{%
{\protect \APACyear {2024}}%
}]{%
sharma2024image}
\APACinsertmetastar {%
sharma2024image}%
\begin{APACrefauthors}%
Sharma, R.%
\BCBT {}\ \BBA {} Kukreja, V.%
\end{APACrefauthors}%
\unskip\
\newblock
\APACrefYearMonthDay{2024}{}{}.
\newblock
{\BBOQ}\APACrefatitle {Image segmentation, classification and recognition methods for comics: A decade systematic literature review} {Image segmentation, classification and recognition methods for comics: A decade systematic literature review}.{\BBCQ}
\newblock
\APACjournalVolNumPages{Engineering Applications of Artificial Intelligence}{131}{}{107715}.
\PrintBackRefs{\CurrentBib}

\bibitem [\protect \citeauthoryear {%
Takahashi%
\ \BBA {} Kondo%
}{%
Takahashi%
\ \BBA {} Kondo%
}{%
{\protect \APACyear {2023}}%
}]{%
takahashi2023textbook}
\APACinsertmetastar {%
takahashi2023textbook}%
\begin{APACrefauthors}%
Takahashi, M.%
\BCBT {}\ \BBA {} Kondo, T.%
\end{APACrefauthors}%
\unskip\
\newblock
\APACrefYearMonthDay{2023}{}{}.
\newblock
{\BBOQ}\APACrefatitle {What is a Textbook Layout that Everyone Understands?: Towards a Comprehensive Understanding of Text and Supplementary Information} {What is a textbook layout that everyone understands?: Towards a comprehensive understanding of text and supplementary information}.{\BBCQ}
\newblock
\BIn{} \APACrefbtitle {Proceedings of the Annual Meeting of the Cognitive Science Society} {Proceedings of the annual meeting of the cognitive science society}\ (\BVOL~45).
\PrintBackRefs{\CurrentBib}

\bibitem [\protect \citeauthoryear {%
T{\"u}chler%
, Zarina%
\BCBL {}\ \BBA {} Skilters%
}{%
T{\"u}chler%
\ \protect \BOthers {.}}{%
{\protect \APACyear {2021}}%
}]{%
tuchler2021impact}
\APACinsertmetastar {%
tuchler2021impact}%
\begin{APACrefauthors}%
T{\"u}chler, A\BPBI F.%
, Zarina, L.%
\BCBL {}\ \BBA {} Skilters, J.%
\end{APACrefauthors}%
\unskip\
\newblock
\APACrefYearMonthDay{2021}{}{}.
\newblock
{\BBOQ}\APACrefatitle {The impact of interface alignment structure on aesthetic appreciation and usability rating} {The impact of interface alignment structure on aesthetic appreciation and usability rating}.{\BBCQ}
\newblock
\BIn{} \APACrefbtitle {Proceedings of the Annual Meeting of the Cognitive Science Society} {Proceedings of the annual meeting of the cognitive science society}\ (\BVOL~43).
\PrintBackRefs{\CurrentBib}

\bibitem [\protect \citeauthoryear {%
Wi%
, Jang%
\BCBL {}\ \BBA {} Kim%
}{%
Wi%
\ \protect \BOthers {.}}{%
{\protect \APACyear {2020}}%
}]{%
wi}
\APACinsertmetastar {%
wi}%
\begin{APACrefauthors}%
Wi, J\BPBI A.%
, Jang, S.%
\BCBL {}\ \BBA {} Kim, Y.%
\end{APACrefauthors}%
\unskip\
\newblock
\APACrefYearMonthDay{2020}{}{}.
\newblock
{\BBOQ}\APACrefatitle {Poster-Based Multiple Movie Genre Classification Using Inter-Channel Features} {Poster-based multiple movie genre classification using inter-channel features}.{\BBCQ}
\newblock
\APACjournalVolNumPages{IEEE Access}{8}{}{66615-66624}.
\PrintBackRefs{\CurrentBib}

\bibitem [\protect \citeauthoryear {%
Young-Min%
}{%
Young-Min%
}{%
{\protect \APACyear {2019}}%
}]{%
young2019feature}
\APACinsertmetastar {%
young2019feature}%
\begin{APACrefauthors}%
Young-Min, K.%
\end{APACrefauthors}%
\unskip\
\newblock
\APACrefYearMonthDay{2019}{}{}.
\newblock
{\BBOQ}\APACrefatitle {Feature visualization in comic artist classification using deep neural networks} {Feature visualization in comic artist classification using deep neural networks}.{\BBCQ}
\newblock
\APACjournalVolNumPages{Journal of Big Data}{6}{1}{56}.
\PrintBackRefs{\CurrentBib}

\end{thebibliography}

\end{document}